\documentclass[letterpaper]{article} % DO NOT CHANGE THIS
\usepackage{aaai2026}  % DO NOT CHANGE THIS
\usepackage{times}  % DO NOT CHANGE THIS
\usepackage{helvet}  % DO NOT CHANGE THIS
\usepackage{courier}  % DO NOT CHANGE THIS
\usepackage[hyphens]{url}  % DO NOT CHANGE THIS
\usepackage{graphicx} % DO NOT CHANGE THIS
\usepackage{natbib}  % DO NOT CHANGE THIS AND DO NOT ADD ANY OPTIONS TO IT
\usepackage{caption} % DO NOT CHANGE THIS AND DO NOT ADD ANY OPTIONS TO IT
\usepackage{adjustbox}
\usepackage{algorithm}
\usepackage{algorithmic}

\usepackage{amsmath}
\usepackage{amssymb}
\usepackage{bm}
\usepackage{booktabs}

\usepackage{newfloat}
\usepackage{listings}
\DeclareCaptionStyle{ruled}{labelfont=normalfont,labelsep=colon,strut=off} % DO NOT CHANGE THIS
\floatstyle{ruled}
\newfloat{listing}{tb}{lst}{}
\floatname{listing}{Listing}
\title{CoCo-MILP: Inter-Variable Contrastive and \\ Intra-Constraint Competitive MILP Solution Prediction}
\author{
    Tianle Pu\textsuperscript{\rm 1,\rm 2}\equalcontrib,
    Jianing Li\textsuperscript{\rm 1,\rm 2}\equalcontrib,
    Yingying Gao\textsuperscript{\rm 1}\equalcontrib,
    Shixuan Liu\textsuperscript{\rm 3,\rm 1,\rm 2},
    Zijie Geng\textsuperscript{\rm 4},
    Haoyang Liu\textsuperscript{\rm 4}, \\
    Chao Chen\textsuperscript{\rm 1,\rm 2},
    Changjun Fan\textsuperscript{\rm 1,\rm 2}\thanks{Corresponding author.}
}
\affiliations{
    \textsuperscript{\rm 1}College of Systems Engineering, National University of Defense Technology\\
    \textsuperscript{\rm 2}Laboratory for Big Data and Decision, National University of Defense Technology\\
    \textsuperscript{\rm 3}College of Computer Science and Technology, National University of Defense Technology\\
    \textsuperscript{\rm 4}MoE Key Laboratory of Brain-inspired Intelligent Perception and Cognition, University of Science and Technology of China\\
    \{putl22, lijianing24, gaoyingying21, liushixuan, chenc1997, fanchangjun\}@nudt.edu.cn, \\ \{zijiegeng, dgyoung\}@mail.ustc.edu.cn

}

\usepackage{bibentry}
\begin{document}

\maketitle

\begin{abstract}
Mixed-Integer Linear Programming (MILP) is a cornerstone of combinatorial optimization, yet solving large-scale instances remains a significant computational challenge. 
Recently, Graph Neural Networks (GNNs) have shown promise in accelerating MILP solvers by predicting high-quality solutions.
However, we identify that existing methods misalign with the intrinsic structure of MILP problems at two levels.
At the leaning objective level, the Binary Cross-Entropy (BCE) loss treats variables independently, neglecting their relative priority and yielding plausible logits.
At the model architecture level, standard GNN message passing inherently smooths the representations across variables, missing the natural competitive relationships within constraints.
To address these challenges, we propose \textbf{CoCo-MILP}, which explicitly models inter-variable \underline{\bf Co}ntrast and intra-constraint \underline{\bf Co}mpetition for advanced MILP solution prediction.
At the objective level, CoCo-MILP introduces the Inter-Variable Contrastive Loss (VCL), which explicitly maximizes the embedding margin between variables assigned one versus zero.
At the architectural level, we design an Intra-Constraint Competitive GNN layer that, instead of homogenizing features, learns to differentiate representations of competing variables within a constraint, capturing their exclusionary nature.
Experimental results on standard benchmarks demonstrate that CoCo-MILP significantly outperforms existing learning-based approaches, reducing the solution gap by up to 68.12\% compared to traditional solvers. Our code is available at https://github.com/happypu326/CoCo-MILP.
\end{abstract}


\section{Introduction}
Mixed-Integer Linear Programming (MILP) is a cornerstone of combinatorial optimization, with diverse applications in the world \citep{fan2020finding, fan2023searching, liu2024inductive, pu2024solving, pu2024exploratory, 2025environment, symmap, wangaccelerating}.
Despite its expressive power, solving MILP instances is fundamentally challenging due to their NP-hard nature.
To tackle this, extensive research has led to the development of sophisticated solvers like SCIP \citep{scip} and Gurobi \citep{gurobi}, which are primarily based on Branch-and-Bound (B\&B) and Branch-and-Cut (B\&C) algorithms \citep{BnB2010, BnC2002}.
Although these solvers are meticulously enhanced with various heuristics, the significant computational burden for large-scale instances remains, motivating the search for new paradigms to accelerate the discovery of high-quality solutions.

\begin{figure}
    \centering
    \includegraphics[width=0.95\linewidth]{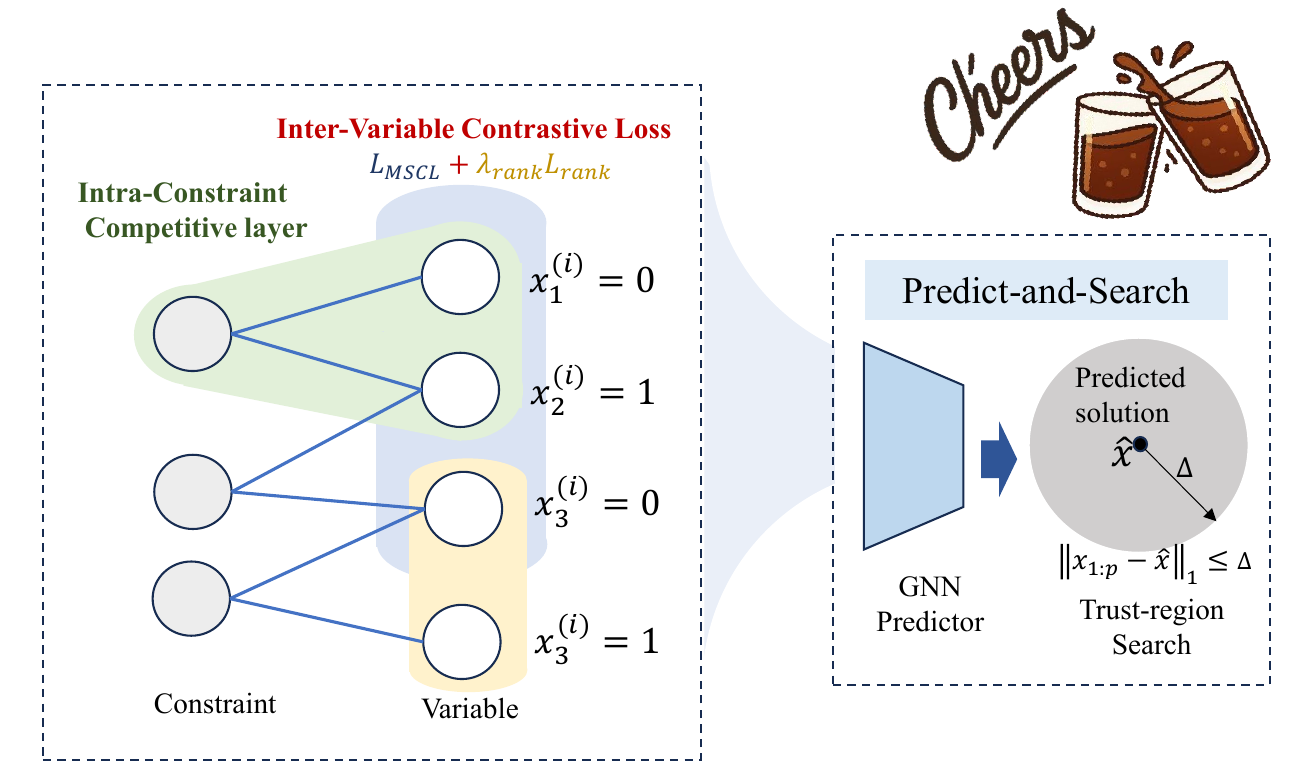}
    \caption{Illustration of CoCo-MILP. We aim to improve the quality of the predicted variables for MILP. The main contribution of our work is the proposed inter-variable contrastive loss and intra-constraint competitive layer for the GNN predictor in the predict-and-search framework.}
    \label{fig:placeholder}
\end{figure}

A promising paradigm has recently emerged to directly predict high-quality primal solutions using machine learning, particularly Graph Neural Networks (GNNs) \citep{learn2branch2019, hem2023}.
These methods typically frame the task in a supervised learning context: by representing MILP instances as graphs, a GNN is trained to map problem structures to variable assignments, using near-optimal solutions generated by traditional solvers as training labels \citep{PS2023, ConPS, liu2025apollomilp}.
This data-driven approach allows the model to learn and exploit structural patterns from past instances to make rapid predictions.
Pioneering works, such as Neural Diving \citep{neuraldiving2020} and the broader predict-and-search strategies \citep{PS2023}, have shown that leveraging these predictions can significantly accelerate the solving process.

However, despite this considerable potential, the efficacy of these learning-based heuristics is fundamentally limited by their underlying design choices. We identify that the standard GNN architectures and learning objectives are not fully aligned with the combinatorial nature of MILP, which manifests as two critical misalignments. First, at the learning objective level, the prevalent use of the Binary Cross-Entropy (BCE) loss function frames the task as a set of independent binary classifications, one for each variable. This formulation inherently neglects the relative priority among variables, which is critical for constructing a globally optimal solution, and consequently yields plausible but poorly separated logits. Second, at the model architecture level, standard GNN message-passing mechanisms inherently smooth representations, making connected nodes more similar. In the context of MILP, however, variables sharing a constraint (e.g., in $\sum_{i}x_i\le 1$) are often mutually competitive \citep{chen2022representing, chen2024expressive, chen2024rethinking, zhang2024expressive}. The smoothing nature of standard GNNs thus works directly against the goal of capturing these crucial competitive relationships, effectively masking the problem's underlying structure.

To address these challenges, we propose CoCo-MILP, a novel framework designed to explicitly model inter-variable Contrast and intra-constraint Competition. To rectify the objective-level misalignment, CoCo-MILP introduces an Inter-Variable Contrastive Loss (VCL), which moves beyond pointwise accuracy to directly optimize the relative ordering of variables by maximizing the embedding margin between variables assigned one versus zero. In parallel, to address the architectural misalignment, we design an Intra-Constraint Competitive GNN layer. This layer is engineered to differentiate, rather than homogenize, the representations of competing variables within a constraint, enabling the model to effectively learn their exclusionary nature.

\section{Preliminaries}
\subsection{Mixed-Integer Linear Programming}
\label{ssec:MILP}
A standard Mixed-Integer Linear Programming (MILP) instance $\mathcal{I}$ is defined as:
\begin{equation}
\min_{\bm{x} \in \mathbb{Z}^p \times \mathbb{R}^{n-p}} \left\{ \bm{c}^\top \bm{x} \;\middle|\; \bm{A} \bm{x} \le \mathbf{b},\; \bm{l} \le \bm{x} \le \bm{u} \right\},
\end{equation}
where \( \bm{x} \in \mathbb{R}^n \) denotes the decision variables, with the first $p$ entries being integer and the remaining $n-p$ continuous.
The vector $\bm{c} \in \mathbb{R}^n $ denotes the coefficients of the linear objective, the constraints are defined by the matrix \( \bm{A} \in \mathbb{R}^{m \times n} \) and the right-hand side vector \( \bm{b} \in \mathbb{R}^m \), and the variable bounds are given by $\bm{l}\in (\mathbb{R}\cup\{-\infty\})^n$ and $\bm{u}=(\mathbb{R}\cup \{+\infty\})^n$.
Without loss of generality, we focus on binary integer variables, i.e., we assume that $\bm{x}\in\{0,1\}^{p}\times\mathbb{R}^{n-p}$.
General integer variables can be handled via standard preprocessing techniques~\cite{nair2020solving}.
Finding the optimal solution that optimizes the objective function is NP-hard, making large-scale instances computationally challenging for exact solvers.

\subsection{Related Work}
\label{sec:related-work}
In recent years, machine learning techniques have seen widespread use in accelerating the solution of MILPs \citep{li2024machine}. These research efforts primarily follow four key directions. One focus is on generating new data to support solver advancement \citep{g2milp2023, liu2024milpstudio}. Another critical direction involves enhancing key modules within solvers including variable selection \citep{learn2branch2019, rl4branch2022, kuang2024rethinking}, node selection \citep{he2014learning, nodecompare, zhang2025learning}, cutting plane selection \citep{hem2023, separator2023, yedynamic}, and large neighborhood search \citep{LNS2021, LNS2023, song2020lns, wu2021lns, ye23GNNGBDT} A third area of exploration centers on predicting high-quality solutions to enable warm-starting of solvers, as demonstrated in studies such as \citep{nair2020solving, PS2023, ConPS, geng2025differentiable, liu2025apollomilp}. Additionally, efforts have been directed at improving the generalization ability of learning-based models \citep{liu2023promotinggeneralizationexactsolvers, ye23GNNGBDT, ye2024lmip, 2025rome}.
We also observe that as Large Language Models (LLMs) grow in popularity, a range of LLM-based methods have appeared, and one notable application among these is LLM-driven optimization modeling \citep{jiang2024llmopt, liuoptitree}. Finally, we provide details on related works in ML4CO and contrastive learning, which are included in Appendix B.

% Recent work on learning-based MILP solvers can be roughly classified into two categories.
% The first line of research integrates machine learning to enhance the solving efficiency of conventional solvers.
% Researchers utilize a learned model to replace some key modules in the solvers, including variable selection \citep{learn2branch2019, rl4branch2022, kuang2024rethinking}, node selection \citep{he2014learning, nodecompare, zhang2025learning}, cutting plane selection \citep{hem2023, separator2023} and large neighborhood search \citep{LNS2021, LNS2023, song2020lns, wu2021lns, ye23GNNGBDT}.
% Another line of research focuses on leveraging learning-based models to predict an initial solution for MILPs.
% \citet{nair2020solving} proposes the first solution prediction framework. 
% Subsequently, \citet{PS2023} proposes the predict-and-search (PS) framework, which introduces a trust region search for better feasibility and improved solution quality. 
% To further enhance the prediction quality, \citet{ConPS} employs contrastive learning to train the solution prediction network.
% To improve the search efficiency after solution prediction, \citet{ye23GNNGBDT, ye2024lmip} predict an initial solution and design sophisticated neighborhood search heuristics. Additionally, we provide a comprehensive survey of related work in ML4CO and contrastive learning in SI 2.

\subsection{Predict-and-Search Framework for MILPs}
\label{ssec:PS}
We can encode each MILP instance as a bipartite graph \( \mathcal{G} = (\mathcal{W} \cup \mathcal{V}, \mathcal{E}) \), where \( \mathcal{W} \) and \( \mathcal{V} \) denote the sets of constraint and variable nodes, respectively, and the edge set $\mathcal{E}$ corresponds to non-zero entries in \( \bm{A} \).
Each node and edge is associated with a set of features derived from problem coefficients and structural attributes.
Such a bipartite graph can completely describe a MILP instance, enabling GNNs to process the instances and predict their solutions.

We adopt the PS paradigm to approximate the solution distribution of a given MILP.
Specifically, the distribution is defined via an energy function that assigns lower energy to high-quality feasible solutions and infinite energy to infeasible ones:
\begin{equation}
\begin{aligned}
    & p(\bm{x}\mid\mathcal{I}) = \frac{\exp(-E(\bm{x}\mid \mathcal{I}))}{\sum_{\bm{x}'}\exp(-E(\bm{x}'\mid \mathcal{I}))},\\
    & \text{where}\quad
    E(\bm{x}\mid \mathcal{I})=
    \begin{cases}
        \bm{c}^\top \bm{x}, & \text{if }\bm{x} \text{ is feasible,}\\
        +\infty,& \text{otherwise.}
    \end{cases}
\end{aligned}
\end{equation}
Our goal is to learning  distribution $p_{\bm{\theta}}(\bm{x}|\mathcal{I})$ for a given instance $\mathcal{I}$.
To make learning tractable, we assume a fully factorized solution distribution over the binary variables, i.e., $p_{\bm{\theta}}(\bm{x}|\mathcal{I})=\prod_{i=1}^{p}p_{\bm{\theta}}(x_i|\mathcal{I})$, where $p_{\bm{\theta}}(x_i|\mathcal{I})$ denotes the predicted marginal probability for variable $x_i$.
To do so, we use a GNN model to output a $p$-dimension vector $\hat{\bm{x}}=\bm{f}_{\bm{\theta}}(\mathcal{I})=(\hat{x}_1,\cdots,\hat{x}_p)^\top\in [0,1]^p$, where $\hat{x}_j=p_{\bm{\theta}}(x_j=1|\mathcal{I})$.
To train the model, we use a weighted set of feasible solutions $\{\bm{x}^{(i)}\}_{i=1}^N$ as supervised signals, where each solution is assigned a weight \( w_i \propto \exp(-\bm{c}^\top \bm{x}^{(i)}) \).
Let $x^{(i)}_j=p_{\bm{\theta}}(x^{(i)}_j=1|\mathcal{I})$.
Then, the training loss function is a binary cross-entropy loss defined as:
\begin{equation}
\label{equ: bce}
\begin{aligned}
&\mathcal{L}_{\text{BCE}}(\bm{\theta}|\mathcal{I}) =  \sum_{i=1}^{N}w_i\cdot\mathcal{L}_{\text{BCE}}(\bm{\theta}|\mathcal{I},\bm{x}^{(i)})\\
=&\sum_{i=1}^{N}w_i\cdot\sum_{j=1}^p -\left[ x_j^{(i)} \log \hat{x}_j + (1 - x_j^{(i)}) \log(1 - \hat{x}_j) \right].
\end{aligned}
\end{equation}
At inference time, the GNN model outputs a predicted marginal $\hat{\bm{x}}\in[0,1]^p$.
A standard MILP solver (e.g., Gurobi or SCIP) is then used to search for a feasible solution in a local neighborhood around $\hat{\bm{x}}$ by solving the following trust region problem:
\begin{equation}
\min_{\bm{x} \in \mathbb{Z}^p \times \mathbb{R}^{n-p}} \left\{ \bm{c}^\top \bm{x} \;\middle|\; \bm{A} \bm{x} \le \mathbf{b},\; \bm{l} \le \bm{x} \le \bm{u},\bm{x}_{1:p}\in \mathcal{B}(\hat{\bm{x}},\Delta) \right\},
\end{equation}
where the trust region $\mathcal{B}(\hat{\bm{x}},\Delta):=\{\bm{x}\in\mathbb{R}^n:\|\bm{x}_{1:p}-\hat{\bm{x}}\|_1\le \Delta\}$ constrains the solver to remain close to the predicted binary configuration.

\section{Motivation}
\label{sec:motivation}
In this section, we motivate our approach by highlighting two fundamental limitations of conventional methods for predicting MILP solutions: (1) their tendency to produce ambiguous logit scores for variables, and (2) their failure to capture the competitive dynamics within constraints. We also provide an extensive motivation analysis from both operations research and application perspectives in Appendix A.

\subsection{Inter-Variable Logits Ambiguity}
A primary goal of a prediction model is to provide a clear signal to a solver about which variables to prioritize. However, we observe that standard models often fail to do so.

\begin{figure}[h]
    \centering
    \includegraphics[width=0.95\linewidth]{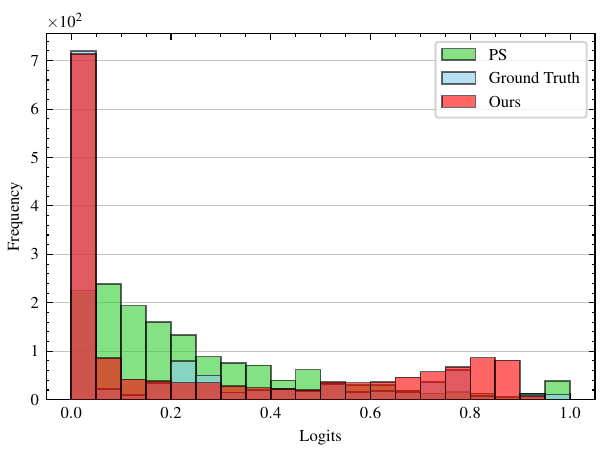}
    \caption{Distribution of Logits from different methods.}
    \label{fig:motivation1}
\end{figure}

\paragraph{Logits Distribution Visualization.}
We begin by visualizing the logit distributions from a baseline GNN model trained with the standard Binary Cross-Entropy (BCE) loss.
As shown in Figure~\ref{fig:motivation1}, the logits produced by PS are poorly separated.
The model assigns similar scores to many variables that should be distinguished, failing to establish a clear decision boundary for reliable variable ranking.
In contrast, the logits produced by our method exhibit a much clearer separation, aligning better with the ground truth.

\begin{figure}[h]
    \centering
    \includegraphics[width=0.95\linewidth]{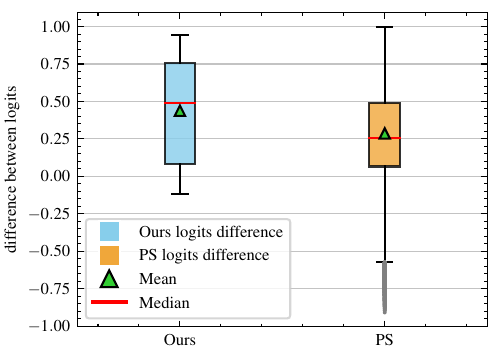}
    \caption{Distribution of the difference between logits corresponding to ground truth 1 and 0 from different methods.}
    \label{fig:motivation2}
\end{figure}

\paragraph{Quantifying Logit Separability.}
To quantify this ambiguity, we measure the model's ability to rank variables correctly. For a given instance, we sample a large number of positive-negative variable pairs $(v_i, v_j)$ from a ground truth solution, where $x_i =1$ and $x_j =0$.
We then compute the difference between the predicted scores, $\Delta_{ij} = \hat{x}_i - \hat{x}_j$, where $\hat{x}_i$ and $\hat{x}_j$ are the predictions for variables $v_i$ and $v_j$, respectively.
A positive difference signifies a correct relative ranking, while a negative one indicates an error.
The statistics of these differences in Figure~\ref{fig:motivation2} reveal that for the baseline model, a substantial portion of the distribution falls into the negative region. This quantitatively confirms that the model is frequently uncertain or even incorrect about the relative importance of variable pairs, underscoring the need for better logit differentiation.

\paragraph{Analysis.}
The standard training objective for MILP solution prediction is the Binary Cross-Entropy (BCE) loss.
This objective encourages the model to independently match the marginal probability for each variable.
However, a high-quality solution for a MILP problem depends not on individual accuracies, but on the correct \textit{relative ordering} of variables.
A solver does not need to know the absolute probability of $x_i=1$; it needs to know whether $x_i$ is a \textit{better} candidate than $x_j$.
A model trained with BCE can be ``correct on average'' yet fail to create a decisive separation between high- and low-priority variables.
This results in ambiguous logits that offer a weak and indecisive signal to the solver, motivating our design of a new loss function that explicitly prioritizes comparative ranking.

\subsection{Neglect of Intra-Constraint Competitions}
\begin{figure}[h]
    \centering
    \includegraphics[width=0.95\linewidth]{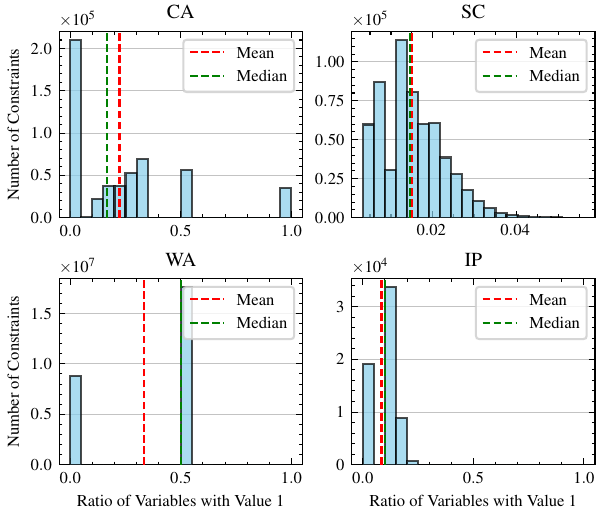}
    \caption{Ratio of variables with value $1$ from different problems. For each problem category, we traverse through all constraints to count the number of variables taking the value 1 in the ground truth and plot the resulting histogram.}
    \label{fig:motivation3}
\end{figure}

A second critical limitation of existing methods is their neglect of the competitive relationships inherent in constraints.
\paragraph{Prevalence of Competition in Constraints.}
Many constraints, such as the set-packing constraint $\sum_{i \in S} x_i \le 1$, enforce mutual exclusivity, where only one or a few variables can be simultaneously active (i.e., set to 1).
To demonstrate the prevalence of this structure, we analyze the activation ratio of constraints across standard benchmark datasets, defined as the proportion of active variables within a constraint in a ground-truth solution.
Figure~\ref{fig:motivation3} plots the distribution of these activation ratios. The plot reveals that for the vast majority of constraints, the activation ratio is extremely low.
This confirms that a "winner-takes-few" or even "winner-takes-one" dynamic is a fundamental and widespread property of MILP instances.

\begin{figure}[h]
    \centering
    \includegraphics[width=0.95\linewidth]{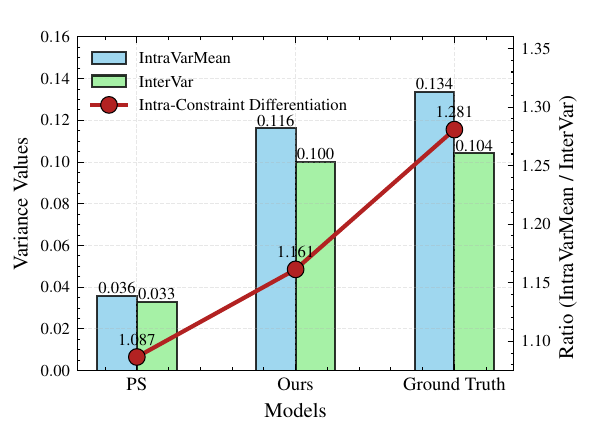}
    \caption{Distribution of logit variance per constraint. IntraVarMean first calculates the variance of logits within each constraint and then computes their average; InterVar calculates the variance across all logits; Ratio represents the ratio of IntraVarMean to InterVar.}
    \label{fig:motivation4}
\end{figure}

\paragraph{Insufficient Differentiation within Constraints.}
We next investigate whether baseline models can capture this competitive structure. An effective model should differentiate between variables within a constraint, assigning high scores to the few ``winners" and low scores to the many ``losers." We measure this differentiation by calculating the variance of logits among all variables sharing a constraint. Figure~\ref{fig:motivation4} compares the distribution of this intra-constraint logit variance for the baseline model against our method. The baseline model's variance is predominantly low, indicating that it assigns indiscriminately similar scores to competing variables. In contrast, our method produces significantly higher variance, demonstrating a superior ability to distinguish between winners and losers within a constraint.

\paragraph{Analysis.}
This failure stems from the core mechanism of the GNNs commonly used for prediction. Standard GNN message passing is designed to smooth and aggregate features among neighboring nodes.
While effective for tasks like node classification, this mechanism is fundamentally ill-suited for modeling competition.
By averaging information from neighbors within a constraint, GNNs tend to homogenize the representations of competing variables. This process inherently masks their exclusionary relationships, effectively treating them as collaborators rather than the competitors they are.

\section{Methodology}
\label{sec:methodology}

Motivated by the observations in Section~\ref{sec:motivation}, we propose \textbf{CoCo-MILP}, a novel framework that explicitly models inter-variable \underline{\bf Co}ntrast and intra-constraint \underline{\bf Co}mpetition for advanced MILP solution prediction.

\subsection{Framework Overview}
\label{ssec:overview}
The CoCo-MILP framework is built upon the PS framework, as introduced in Section~\ref{ssec:PS}.
In the framework, a MILP instance is represented by a variable-constraint bipartite graph, and a GNN is trained to predict the values of binary variables.
Our analysis in Section~\ref{sec:motivation} revealed two critical weaknesses in the standard PS approach: the ambiguous variable logits and honogenized representations.
We attribute these issues to the inadequate learning objective and the GNN architecture neglecting variable competitions.

CoCo-MILP addresses the aforementioned issues with two core innovations.
In Section~\ref{ssec:inter-variable-contrastive-loss}, we introduce the Inter-Variable Contrastive Loss (VCL), which replaces the standard BCE loss to learn more discriminative variable logits by focusing on their relative contrast.
Then, in Section~\ref{ssec:intra-constraint-competitive-gnn}, we introduce the Intra-Constraint Competitive (ICC) GNN layer.
This layer explicitly models the competitive relationships among variables within shared constraints, thereby counteracting the feature-smoothing effect of standard message passing.

\subsection{Inter-Variable Contrastive Loss (VCL)}
\label{ssec:inter-variable-contrastive-loss}
The conventional Binary Cross-Entropy (BCE) loss function treats the prediction for each variable as an independent classification task.
This approach disregards the crucial relational context between variables, leading to the ambiguous logits demonstrated in Section~\ref{sec:motivation}.
To address this, we propose the Inter-Variable Contrastive Loss (VCL), which fundamentally shifts the learning objective from pointwise accuracy to learning discriminative logits.
The goal of VCL is to structure the embedding space such that the logits of variables with a ground-truth value of one are systematically and confidently larger than those with a ground-truth value of zero.

Our VCL is a composite objective composed of two complementary loss functions: a global multi-sample contrastive loss and a pairwise ranking loss.
For a given MILP instance $\mathcal{I}$, let our model with parameters $\bm{\theta}$ produce a vector of logits $\bm{z}$ and the corresponding prediction $\hat{\bm{x}}$.
For a ground-truth solution $\bm{x}^{(i)}$, we partition the set of binary variables $\mathcal{V}_0$ into a positive set $\mathcal{V}_{+} = \{v_i \mid x_i^{(i)}=1\}$ and a negative set $\mathcal{V}_{-} = \{v_j \mid x_j^{(i)}=0\}$.

First, we introduce the Multi-Sample Contrastive Loss (MSCL), which is inspired by InfoNCE, which adopts a global perspective.
It encourages the logits of all positive samples to be collectively larger than the logits of all other variables in the instance.
This is formulated as:
\begin{equation}
\label{eq:mscl}
\mathcal{L}_{\text{MSCL}}(\bm{\theta} \mid \mathcal{I}, \mathbf{x}^{(i)}) = - \log\frac{\sum_{v_i \in \mathcal{V}_{+}}\exp(z_i/\tau)}{\sum_{v_k \in \mathcal{V}_{0}}\exp(z_k/\tau)},
\end{equation}
where $\tau$ is a temperature hyperparameter.
This loss pushes the entire group of positive variables away from the entire pool of variables, promoting a coarse-grained separation.

Second, to enforce a more fine-grained separation, we employ a Pairwise Ranking Loss.
This loss focuses on the local relationship between every individual positive-negative pair, ensuring a strict margin between their logits.
It is defined as:
\begin{equation}
\label{eq:rank}
\begin{aligned}
&\mathcal{L}_{\text{rank}}(\bm{\theta} \mid \mathcal{I}, \mathbf{x}^{(i)}) \\
=& \frac{1}{|\mathcal{V}_{+}| |\mathcal{V}_{-}|} \sum_{v_i \in \mathcal{V}_{+}, v_j \in \mathcal{V}_{-}} \max(0, \gamma - (z_i - z_j)),
\end{aligned}
\end{equation}
where $\gamma > 0$ is a predefined margin. This loss penalizes any pair $(v_i, v_j)$ where the logit of the positive variable $z_i$ does not exceed the logit of the negative variable $z_j$ by at least $\gamma$.

The final Inter-Variable Contrastive Loss (VCL) is the weighted sum of these two components:
\begin{equation}
\label{eq:vcl}
\begin{aligned}
&\mathcal{L}_{\text{VCL}}(\bm{\theta} \mid \mathcal{I})=\sum_{i=1}^{N}w_i\cdot\mathcal{L}_{\text{VCL}}(\bm{\theta} \mid \mathcal{I}, \mathbf{x}^{(i)}) \\
= &\sum_{i=1}^{N}w_i\cdot\left[\mathcal{L}_{\text{MSCL}}(\bm{\theta} \mid \mathcal{I}, \mathbf{x}^{(i)}) + \lambda_{\text{rank}}\cdot \mathcal{L}_{\text{rank}}(\bm{\theta} \mid \mathcal{I}, \mathbf{x}^{(i)})\right],
\end{aligned}
\end{equation}
where $w_i$ are weights for each solution and $\lambda_{\text{rank}}$ is a coefficient balancing the two objectives.
Unlike BCE, which evaluates each variable in isolation, VCL is relational; its gradient for any single variable depends on the logits of all other variables.
By combining a global contrastive push with fine-grained pairwise ranking, VCL produces a much clearer and more robust ranking signal, which is better aligned with the needs of downstream heuristics and solvers.

\subsection{Intra-Constraint Competitive GNN}
\label{ssec:intra-constraint-competitive-gnn}
Standard GNN message passing tends to smooth features across connected nodes, an effect that is counterproductive in the MILP context.
As discussed in Section~\ref{sec:motivation}, variables sharing a constraint are not collaborators but competitors.
An effective GNN architecture should therefore not homogenize its representations but instead \textit{differentiate} them to highlight the most promising candidates for inclusion in a solution.

To achieve this, we propose that a variable's representation should be contextualized by its local competition. Instead of learning an absolute embedding, we aim to learn how each variable's features \textit{deviate} from the average features of its direct competitors within each constraint. This relative representation is inherently more discriminative.

We implement this principle with our Intra-Constraint Competitive (ICC) layer. The ICC layer achieves differentiation through a simple yet effective normalization process that follows each standard GNN message-passing update.
Let $\bm{h}_j^{(l)}$ be the embedding of a variable $v_j$ after the $l^\text{th}$ message-passing layer.
The ICC mechanism proceeds in three steps.

\paragraph{Aggregate Competitor Features.}
For each constraint $c_k$, we compute an aggregated message by averaging the embeddings of all variables participating in it:
\begin{equation}
\overline{\bm{h}}_k^{(l)} \gets \frac{1}{|\mathcal{N}(c_k)|} \sum_{v_j \in \mathcal{N}(c_k)} \bm{h}_j^{(l)},
\end{equation}
where $\mathcal{N}(c_k)$ denotes the set of variable nodes connected to constraint $c_k$.
The vector $\overline{\bm{h}}_k^{(l)}$ represents the ``average competitor'' within that constraint's neighborhood.

\paragraph{Propagate Competitive Context.}
The aggregated information is propagated back to the variables.
Each variable $v_j$ aggregates the ``average competitor'' representations from all constraints it participates in:
\begin{equation}
\overline{\bm{h}}_j^{(l)} \gets \frac{1}{|\mathcal{N}(v_j)|} \sum_{c_k \in \mathcal{N}(v_j)} \overline{\bm{h}}_k^{(l)},
\end{equation}
where $\mathcal{N}(v_j)$ is the set of constraint nodes connected to variable $v_j$.
The resulting vector $\overline{\bm{h}}_j$ represents the aggregated features of the average peer group against which variable $v_j$ competes.

\paragraph{Compute Competitive Deviation}
The final update is then performed by subtracting the peer representation from the variable's embedding, scaled by a learnable parameter $\beta$:
\begin{equation}
\label{eq:competitive_update}
\bm{h}_j^{(l)} \gets \bm{h}_j^{(l)} - \beta \cdot \overline{\bm{h}}_j^{(l)}
\end{equation}

This mechanism explicitly calculates the deviation of a variable's feature from its competitive baselines.
A variable whose embedding is highly distinct from its peers will produce a resultant vector with a large magnitude, signaling it as a strong, standout candidate.
Conversely, a variable that is feature-wise similar to its competitors will have its representation pushed towards the zero vector, diminishing its salience.
This process directly counteracts the feature smoothing of standard GNNs and forces the model to learn embeddings that highlight the ``winners'' within each local competition, producing representations that are far more differentiated and informative for the final prediction task.

\section{Experiments}
\label{sec:experiments}
In this section, we conduct extensive experiments to evaluate the effectiveness of CoCo-MILP. Our approach achieves significant improvements in solving performance on both the synthetic (Section \ref{sec:main_results}) and real-world (Section \ref{sec:realworld_results}) benchmarks, as well as generalization ability (Appendix E.4).

\subsection{Experimental Setup}
\label{section:experiment setup}
\paragraph{Datasets}
We evaluate our framework on four well-established MILP benchmarks frequently used in ML4CO research: Set Covering (SC), Item Placement (IP), Combinatorial Auctions (CA), and Workload Appointment (WA). These benchmarks cover a wide range of combinatorial structures, constraint types, and objective complexities. For SC and CA, we follow standard instance generation procedures from current works \citep{learn2branch2019, hem2023, PS2023, ConPS}, while IP and WA are taken from the more challenging tracks of the NeurIPS 2021 ML4CO competition \citep{ml4co}. Following the experimental setting in \cite{PS2023}, we use 240 instances for training, 60 for validation, and 100 for testing. Please refer to Appendix D.1 for further benchmark details.

\paragraph{Baselines}
We mainly evaluate our method through two standard categories. First, we measure the performance gains by integrating the heuristics learned by CoCo-MILP into two typical solvers, Gurobi \citep{gurobi} and SCIP \citep{scip}. Second, we compare CoCo-MILP against two representative learning-based baselines: Predict-and-Search (PS) \citep{PS2023} and Contrastive Predict-and-Search (ConPS) \citep{ConPS}. Both PS and ConPS follow a two-stage paradigm which first predicts partial solutions and then finds better primal solutions based on the Gurobi or SCIP solvers. ConPS extends PS by incorporating contrastive learning to enhance predictive quality. Neural Diving (ND) \citep{neuraldiving2020} is another related baseline with a similar structure to PS, but we do not include it in our comparison due to its weaker performance reported in PS. Most importantly, we are aware of recent stronger baselines, notably Apollo-MILP \citep{liu2025apollomilp}, which focuses on improving search efficiency without modifying the training process of PS. Therefore, to demonstrate the compatibility of our method, we also integrate CoCo-MILP into Apollo-MILP as a plug-in component and report the results in Appendix E.5. More experimental details can be found in Appendix D.

\paragraph{Metrics}
To evaluate solution quality, we compare the best objective values ($\text{OBJ}$) obtained by each method within a 1000-second time limit on each test instance. To approximate the optimal value, we adopt the setting from \citet{PS2023}, where a single-threaded Gurobi is executed for 3600 seconds and its best result is recorded as the best-known solution (BKS). We then compute the absolute primal gap as $\text{gap}_{\text{abs}} := |\text{OBJ} - \text{BKS}|$, which quantifies how far a solution is from the BKS. A smaller $\text{gap}{\text{abs}}$ reflects higher solution quality under the same time budget.

\paragraph{Training and Inference}
Following the setup in PS, all methods are trained for 1000 epochs on each dataset, with the best model selected based on validation prediction loss. To enhance traditional solvers, PS introduces additional constraints to confine the solution space within a trust region, which is controlled by several key hyperparameters. However, since our benchmarks are more challenging and complex, the hyperparameter settings reported in the original papers result in poor performance. Therefore, we re-tune the hyperparameters for all baselines on our benchmarks. The final hyperparameters are detailed in Appendix D.3. And the extensive analysis of these parameters is presented in Appendix E.2.

\subsection{Main Evaluation} \label{sec:main_results}
\paragraph{Solving Performance}
Table \ref{tab:main-results} presents comprehensive results demonstrating that our method consistently achieves state-of-the-art performance across all benchmarks. On the SC benchmark, our method exactly matches the BKS with a near-zero primal gap, yielding a 93.8\% improvement over vanilla Gurobi. For the industrial-scale IP and WA benchmarks, our method achieves perfect optimality, fully matching BKS values. Even on the more complex CA benchmark, our approach narrows the absolute gap by 37.1\% compared to Gurobi. In comparison with learning-based methods, our model reduces the primal gap by 78.6\% against PS and 63.0\% against ConPS, underscoring the effectiveness of our architecture in enhancing traditional solvers.

\begin{figure*}[t]
\centering
\includegraphics[width=0.90\textwidth]{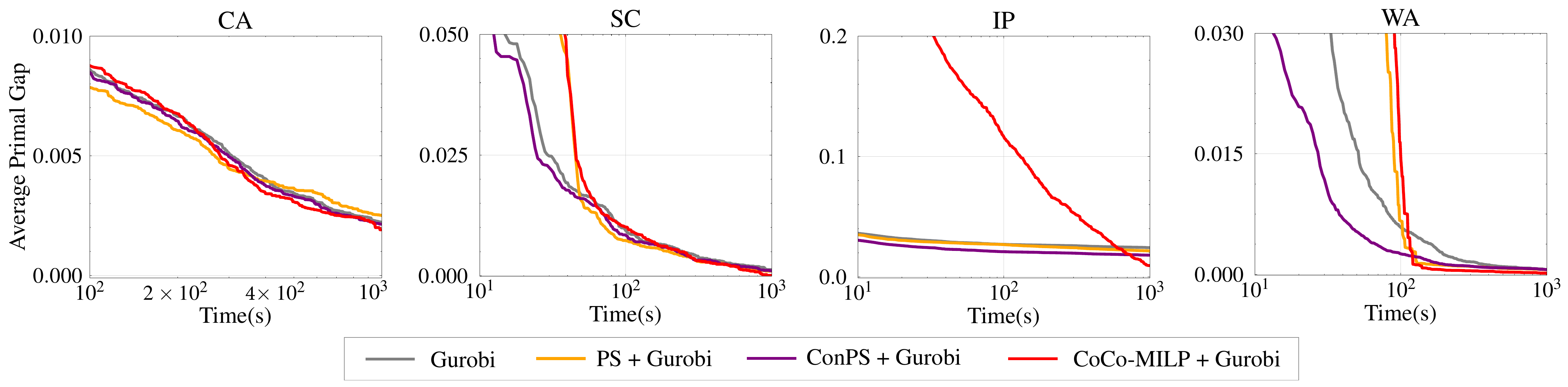}
% \vspace{-0.5cm}
\caption{The average primal gap of each method over 100 test instances as a function of solving time. All methods are implemented using Gurobi, with a time limit of 1000 seconds.
}
% \vspace{-0.25cm}
\label{fig:primal_gap}

\end{figure*}

\paragraph{Primal Gap as a Function of Runtime}
As shown in Figure \ref{fig:primal_gap}, our method demonstrates strong convergence across all benchmarks. Although the initial gap reduction appears less sharp, this is primarily due to the scaling of the x-axis. In practice, CoCo-MILP is able to reach high-quality solutions within the first 100 seconds. This behavior is attributed to the more accurate variable selection in our method. The fixed variables predicted by CoCo-MILP provide a better starting point for the solver, enabling more efficient local search. In contrast, baseline methods often make early decisions based on pointwise training losses, which can misidentify key variables and lead to premature convergence toward suboptimal solutions. The consistently better final objective values achieved by CoCo-MILP further support its advantage in both convergence speed and solution quality. In addition, we provide the results obtained with the SCIP solver for comparison, as presented in Appendix E.3.

\begin{table*}[t]
\centering
\small % 使用较小字体
\caption{Performance comparison on four MILP benchmarks (CA, SC, IP, and WA) under a 1000-second time limit.
‘$\uparrow$’ indicates higher is better; ‘$\downarrow$’ indicates lower is better.
\textbf{Bold} denotes the best result. All improvements are reported as $\text{gap}_{\text{abs}}$ relative to Gurobi.}
\label{tab:main-results}
\resizebox{\textwidth}{!}{ % 自动缩放至栏宽
\begin{tabular}{@{}lcccccccccc@{}}
\toprule
& \multicolumn{2}{c}{CA {\scriptsize(BKS 97524.37)}}  & \multicolumn{2}{c}{SC  {\scriptsize(BKS 125.05)}} & \multicolumn{2}{c}{IP {\scriptsize(BKS 11.16)}} &  \multicolumn{2}{c}{WA {\scriptsize(BKS 703.05)}}\\ 
\cmidrule(lr){2-3}
\cmidrule(lr){4-5}
\cmidrule(lr){6-7}
\cmidrule(lr){8-9}
& Obj $\uparrow$& \hspace{-2mm} $\text{gap}_{\text{abs}}$ $\downarrow$ \hspace{-2mm} & Obj $\downarrow$ \hspace{-2mm} & $\text{gap}_{\text{abs}}$ $\downarrow$ \hspace{-2mm} & Obj $\downarrow$  \hspace{-2mm} & $\text{gap}_{\text{abs}} $ $\downarrow$ \hspace{-2mm} & Obj $\downarrow$ \hspace{-2mm} & $\text{gap}_{\text{abs}}$ $\downarrow$          \\ 
\midrule
Gurobi                 & 97228.93 & 295.44   & 125.21 & 0.16 & 11.43 & 0.27 & 703.47 & 0.42 \\
PS+Gurobi              & 97286.29 & 238.08   & 125.17 & 0.12 & 11.40 & 0.24 & 703.47 & 0.42 \\
ConPS+Gurobi           & 97315.83 & 208.54   & 125.18 & 0.13 & 11.36 & 0.20 & 703.47 & 0.42 \\
CoCo-MILP$+$Gurobi & \textbf{97338.64} & \textbf{185.73} & \textbf{125.06} & \textbf{0.01} & \textbf{11.26} & \textbf{0.10} & \textbf{703.14} & \textbf{0.09}  \\
\midrule
Improvement & & 37.1\% & & 93.8\% & & 63.0\% & & 78.6\%\\
\bottomrule
\end{tabular}}
\end{table*}

\subsection{Evaluation on Real-world Benchmark}\label{sec:realworld_results}
To further evaluate CoCo-MILP’s applicability, we conduct experiments on the MIPLIB dataset \citep{miplib2021}. Following \citep{hem2023, liu2025apollomilp}, we focus on the IIS subset of MIPLIB, which contains six training instances and five testing instances. The complete information about the MIPLIB benchmark, please refer to Appendix D.1. All methods are trained identically, and their performance is reported in Table \ref{tab:miplib_results}. We also evaluate CoCo-MILP on additional challenging subsets of MIPLIB and report the results in Appendix E.1.

\begin{table}[h]
\caption{ The best objectives found by the approaches on each test instance in MIPLIB.
We obtain the \textit{BKS} from the website of MIPLIB. The ML approaches are implemented using Gurobi with a time limit set to 1000 seconds.
}
\vspace{0.25cm}
\centering
\small
%\begin{adjustbox}{width=0.5\textwidth}
{
\begin{tabular}{lcccccc}
\toprule
 &  BKS & Gurobi & PS & ConPS & CoCo-MILP  \\
  \midrule
ex1010-pi & 233.00  & 239.00 & 241.00 & 239.00 & \textbf{237.00} \\
fast0507 & 174.00 & 174.00 & 179.00 & 179.00 & \textbf{174.00} \\
ramos3& 186.00 & 233.00 & 225.00 & 225.00 & \textbf{224.00}\\
scpj4scip& 128.00 & 132.00  & 133.00 & 133.00 & \textbf{131.00}\\
scpl4& 259.00 & 277.00 & 275.00 & 275.00 & \textbf{273.00}\\
\bottomrule
\end{tabular}
%\end{adjustbox}
}
\label{tab:miplib_results}
\end{table}

\subsection{Ablation Study}
To evaluate the contribution of each component in CoCo-MILP, we compare the full model with four ablated variants:
(i) Replacing the VCL loss with $\mathcal{L}_{\text{BCE}}$;
(ii) Removing the ranking terms;
(iii) Removing the MCSL terms;
(iv) Removing the ICC layers in the GNN.
As shown in Table \ref{tab:ablation_study}, omitting either the VCL loss or ICC layers significantly degrades performance, confirming that these design elements play a critical role in CoCo-MILP's effectiveness.

\begin{table}[h]
\centering
\caption{Ablation studies. The ML approaches are implemented using Gurobi with a time limit set to 1000 seconds. `$\uparrow$' indicates that higher is better, and `$\downarrow$' indicates that lower is better. We mark the \textbf{best values} in bold. 
}
\begin{tabular}{@{}ccccc@{}}
\toprule
  & SC  {\scriptsize(BKS 125.05)} & CA {\scriptsize(BKS 97524.37)} \\
\midrule
CoCo-MILP w.i. $\mathcal{L}_{\text{BCE}}$ & 125.25 & 97240.00  \\
CoCo-MILP w/o $\mathcal{L}_{\text{rank}}$ & 125.21 & 97272.41  \\
CoCo-MILP w/o $\mathcal{L}_{\text{MSCL}}$ & 125.19 & 97217.40  \\
CoCo-MILP w/o $\mathcal{L}_{\text{ICC layer}}$ & 125.18 & 97315.96  \\
CoCo-MILP & \textbf{125.06} & \textbf{97338.64} \\
\bottomrule
\end{tabular}
\label{tab:ablation_study}
\end{table}

\section{Conclusion}
In this paper, we summarize the limitations of current learning-based methods and propose CoCo-MILP to address through explicit modeling of inter-variable contrast and intra-constraint competition. Experiments demonstrate that CoCo-MILP outperforms other ML-based approaches, exhibiting promising real-world applicability.

\section{Acknowledgments}
This work was supported by the National Natural Science Foundation of China (NSFC, 72571284, 72421002, 62206303, 62273352), the Hunan Provincial Fund for Distinguished Young Scholars (2025JJ20073), the Science and Technology Innovation Program of Hunan Province (2023RC3009), the Foundation Fund Program of National University of Defense Technology (JS24-05), the Major Science and Technology Projects in Changsha, China (kq2301008), and the Hunan Provincial Innovation Foundation for Postgraduate (XJQY2025044).

\bibliography{aaai2026}

\clearpage
\newpage
\onecolumn % <-- 切换到单栏布局
\appendix
\begin{center}
  \Large \textbf{Appendix}
\end{center}
\vspace{1em} % 这会增加一点标题和正文之间的垂直间距

\section{Extensive Analysis of Motivations}
\paragraph{Inter-Variable Logits Ambiguity}
In many practical combinatorial optimization problems, the items or objects we are interested in are inherently sparse. 
This means the set of variables assigned a value of 1 is significantly smaller than the set of variables assigned a value of 0.
The binary cross-entropy training objective treats the prediction for each variable as an independent binary classification problem.
By focusing on each variable in isolation, the model is not explicitly incentivized to create a large separation between the scores (logits) of positive-class variables and negative-class ones. 
As a result, many variables from the negative set can receive higher scores than variables from the positive set, leading to a significant overlap in their logit distributions and a failure to establish a clear global decision boundary.

\paragraph{Neglect of Intra-Constraint Competitions}
We analyze the cause of the issue. 
The design of standard GNN message-passing mechanisms inherently causes feature smoothing, a process where the representations of connected nodes become more similar after each layer of aggregation. 
In the bipartite graph representation of a MILP, variables that appear in the same constraint are immediate (1-hop) neighbors of that constraint node. 
Consequently, the GNN's smoothing effect is direct and strong, tending to homogenize the representations of these variables.

\paragraph{Explanation under the OR Context} The need to differentiate variables within a constraint is common in combinatorial optimization.
\begin{itemize}
    \item First, this principle of competition is common in diverse real-world problems, such as scheduling, planning, logistics, and network design.
For example, in scheduling, a single machine can only perform one job at a time, making the jobs competitors for that time slot.
In logistics and network design, choosing to build a warehouse in one location or select one delivery route competitively excludes other options.
\item Second, many canonical MILP formulations are built upon constraints that enforce strict differentiation.
For example, one of the most common structures in combinatorial optimization, the set-packing constraint, is explicitly defined as $\sum_{x_i\in S}x_i\le 1$.
The assignment constraints ensure that each worker can only be assigned one task, i.e., $\sum_{j}x_{ij}= 1$.
The knapsack or budget constraints only select a small subset of items in the candidates, i.e., $\sum_iw_ix_i\le W$.
While the specific formulations differ, the underlying principle of competition among variables governed by a single logical or resource-based constraint is a cornerstone of MILP modeling. 
\end{itemize}
The inability of standard GNNs to capture these essential competitive dynamics motivates the development of a specialized architecture, the Intra-Constraint Competitive (ICC) GNN layer, which is explicitly designed to differentiate these competing variable representations.

\section{Related Works}
\paragraph{Machine Learning for MILPs}
Machine learning has been widely applied to accelerate the solving process of MILPs. 
Existing methods can be roughly classified into two categories: machine learning inside an exact solver and machine learning for solution prediction. 
In the first category, researchers apply machine learning models to replace certain heuristics for acceleration. 
These works include learning to branch \cite{learn2branch2019, khalil2022mip, kuang2024rethinking}, learning for cut selection and separation \cite{hem2023, separator2023}, learning to select nodes \cite{nodecompare, he2014learning}, and so on. 
These works have demonstrated significant improvement under an exact solver framework. 
The second category of the works focuses on directly predicting a solution for MILP. 
These methods first use a neural network to predict a high-quality partial solution, and then search in a neighborhood of the partial solution \cite{neuraldiving2020, PS2023, ConPS}. 
The neighborhood search process greatly reduces the search space and thus has gained more and more popularity.

\paragraph{Contrastive Learning}
Contrastive learning has emerged as a powerful technique to learn embeddings that pull similar (positive) samples together and push dissimilar (negative) samples apart. 
Contrastive learning achieves the goal by designing the contrastive functions. 
For pair-wise contrastive learning, triplet loss \cite{chechik2010large, schroff2015facenet} was designed to learn embeddings by ensuring that an anchor sample is closer to a positive sample than to a negative sample by a certain margin. 
The multiple-pair contrastive loss extended this by using multiple negative samples for each positive pair, leading to more efficient training. 
The InfoNCE loss \cite{bachman2019learning,hjelm2018learning,oord2018representation}, a key component in many modern contrastive methods, is derived from noise-contrastive estimation and aims to maximize the mutual information between different views of the data. 
The NT-Xent loss \cite{chen2020simple}, a normalized temperature-scaled cross-entropy loss, is a popular variant that has been widely adopted.
In our work, contrastive learning directly addresses the challenge of ambiguous predictions by directly maximizing the relative difference between variables that should be 1 and 0.

\section{Implementation Details}

\subsection{Graph Representation}
Following prior work~\citep{learn2branch2019, PS2023}, we represent each MILP instance as a weighted bipartite graph $\mathcal{G} = (\mathcal{V} \cup \mathcal{W}, \mathcal{E})$, where $\mathcal{V}$ and $\mathcal{W}$ denote the sets of variable nodes and constraint nodes, respectively, and $\mathcal{E}$ represents the nonzero entries in the constraint matrix.
Each node and edge in the bipartite graph is associated with a set of numerical features derived from MILP coefficients and structural statistics. Our design closely follows the feature definitions used in previous works, and we adopt the same preprocessing pipeline without modification. These features provide rich information for the GNN to exploit the structure of MILP instances effectively.

\subsection{Model Architecture}
Our model builds upon the GNN encoder architecture proposed by \citet{bachman2019learning}, which is also adopted in recent works~\citep{PS2023, geng2025differentiable, liu2025apollomilp}. Given a bipartite graph $\mathcal{G} = (\mathcal{V} \cup \mathcal{W}, \mathcal{E})$, we initialize the node and edge embeddings as:

\begin{equation}
h^{(0)}_{v_i} = \mathrm{MLP}_\theta(v_i), \quad h^{(0)}_{w_j} = \mathrm{MLP}_\theta(w_j), \quad h_{e_{ij}} = \mathrm{MLP}_\theta(e_{ij}).
\end{equation}

Then we apply $K$ layers of half-convolutions:

\begin{equation}
h^{(k+1)}_{w_i} \leftarrow \mathrm{MLP}_\theta\left(h^{(k)}_{w_i}, \sum_{j: e_{ij} \in \mathcal{E}} \mathrm{MLP}_\phi\left(h^{(k)}_{w_i}, h_{e_{ij}}, h^{(k)}_{v_j}\right)\right),
\end{equation}
\begin{equation}
h^{(k+1)}_{v_j} \leftarrow \mathrm{MLP}_\phi\left(h^{(k)}_{v_j}, \sum_{i: e_{ij} \in \mathcal{E}} \mathrm{MLP}_\phi\left(h^{(k+1)}_{w_i}, h_{e_{ij}}, h^{(k)}_{v_j}\right)\right).
\end{equation}

After each message passing layer, we insert our Intra-Constraint Competitive (ICC) module to model exclusion among variables within the same constraint. Specifically, for each variable $v_j$, we compute:

\begin{equation}
\bar{h}^{(k)}_{j} = \frac{1}{|\mathcal{N}(v_j)|} \sum_{c \in \mathcal{N}(v_j)} \left( \frac{1}{|\mathcal{N}(c)|} \sum_{v \in \mathcal{N}(c)} h^{(k)}_v \right),
\end{equation}
\begin{equation}
h^{(k)}_{v_j} \leftarrow h^{(k)}_{v_j} - \beta \cdot \bar{h}^{(k)}_{j},
\end{equation}
where $\beta$ is a learnable scalar.

After $K$ layers, we apply a concatenation-based Jumping Knowledge (JK) mechanism to obtain the final representations:

\begin{equation}
h_{v_i} = \mathrm{MLP}_\theta\left(\mathrm{CONCAT}_{k=0}^{K} h^{(k)}_{v_i}\right),
\end{equation}

\begin{equation}
z_{v_i} = \mathrm{MLP}_\theta(h_{v_i}),
\end{equation}
where $z_{v_i}$ is the predicted logit for variable $v_i$, followed by a sigmoid to obtain the marginal probability.

The ICC module introduces minimal additional overhead, as it only requires one scalar parameter $\beta$ per layer and reuses intermediate embeddings from the base architecture.

\subsection{Implementation of our method and the baselines}
We compare CoCo-MILP against two recent state-of-the-art learning-based approaches: Predict-and-Search (PS)\citep{PS2023} and its contrastive variant ConPS\citep{ConPS}. For PS, we directly adopt the official implementation provided by the authors, including the encoder architecture.
As ConPS does not release its source code, we reproduce its framework based on the algorithmic details in the original paper. To ensure fair comparison, we carefully re-tuned its hyperparameters and verified the reproduced performance against reported results. In our implementation of ConPS, the positive-to-negative sample ratio is set to 1:10, and negative samples are drawn from low-quality feasible solutions.
All experiments are conducted under a consistent computational environment, using an NVIDIA GeForce RTX 3090 GPU and Intel(R) Xeon(R) Gold 5320 CPU (26 cores, 2.20GHz).

\section{Experimental Details}

\subsection{Details of Benchmarks}

\noindent \textbf{Main Evaluation.} 
The CA and SC benchmark instances are constructed based on standard procedures used in prior works. In particular, the CA instances follow the generation algorithm from~\citep{cauctiongen2000}, and the SC instances are derived using the approach proposed in~\citep{setcovergen1980}, as summarized in~\citep{learn2branch2019}. For the IP and WA benchmarks, we adopt the instance set released as part of the NeurIPS 2021 ML4CO competition~\citep{gasse2022machine}. Table~\ref{tab:instance_stat} summarizes the key statistics of all benchmark instances used in our paper.
\begin{table}[h]
\caption{ Statistical information of the benchmarks we used in this paper.}
\centering
\begin{adjustbox}{width=0.85\textwidth}
\begin{tabular}{lccccc}
\toprule
  &  \# Constraint &  \# Variable & \# Binary Variables & \# Continuous Variables & \# Integer Variables \\
  \midrule
CA & 2593  & 1500 & 1500 & 0 & 0\\
SC & 3000 & 5000 & 5000 & 0 & 0\\

IP & 195& 1083& 1050& 33 & 0\\
WA & 64306& 61000& 1000& 60000 & 0\\
\bottomrule
\end{tabular}
\end{adjustbox}
\label{tab:instance_stat}
\end{table}

\noindent \textbf{Generalization.}
To evaluate the cross-scale generalization capacity of our method, we construct a new set of large-scale instances for the CA and SC benchmarks. These instances are synthesized using publicly available generation tools from prior work \citep{learn2branch2019}. The constructed CA problems contain approximately 2,600 constraints and 4,000 variables on average, while the SC problems are built with around 6,000 constraints and 10,000 variables. Compared to the training set, these evaluation instances are significantly larger in both dimensionality and combinatorial complexity, providing a more rigorous test for evaluating model transferability across instance scales.

\noindent \textbf{MIPLIB Subset.}
To evaluate the effectiveness of CoCo-MILP on real-world combinatorial problems, we construct a benchmark based on selected instances from the MIPLIB 2017 dataset \citep{miplib2021}. MIPLIB contains a diverse collection of challenging MILP problems drawn from practical domains such as transportation, energy, logistics, and industrial design. Due to the dataset’s structural heterogeneity, directly training on the full set is often infeasible for learning-based methods.
We adopt an instance similarity-based selection strategy, consistent with prior work such as Apollo-MILP \citep{liu2025apollomilp}. Specifically, similarity is computed using 100 hand-crafted instance features, as introduced in \citet{miplib2021}. Following this protocol, we first utilize the IIS subset from MIPLIB, which consists of eleven real-world instances. Among them, six are used for training—\texttt{glass-sc}, \texttt{iis-glass-cov}, \texttt{5375}, \texttt{214}, \texttt{56133}, and \texttt{iis-hc-cov}—and five are used for testing, namely \texttt{ex1010-pi}, \texttt{fast0507}, \texttt{ramos3}, \texttt{scpj4scip}, and \texttt{scpl4}.
To further challenge the model’s generalization ability, we extend our evaluation to a new set of larger and more difficult MIPLIB instances. For each selected test instance, we construct a training set by identifying approximately five similar instances based on the same feature similarity criteria. All models are trained on these instance-specific training sets and then evaluated on their corresponding target instances.
Table~\ref{table:MIPLIB} summarizes the detailed statistics of the full set of test instances used in our MIPLIB benchmark. The instances span a wide range in terms of size, including both medium- and large-scale problems. Notably, the variable types vary across instances, covering binary, general integer, and continuous variables. This diversity enables us to assess the robustness and versatility of CoCo-MILP across different variable domains. Experimental results (see Appendix D) confirm that CoCo-MILP consistently accelerates solution quality, regardless of variable type.

\begin{table}[h]
\caption{ Statistical information of the instances in the constructed MIPLIB  dataset.}
\centering
\begin{adjustbox}{width=0.9\textwidth}
\begin{tabular}{lcccccc}
\toprule
   & \# Constraint  & \# Variable & \# Binaries  & \# Integers  & \# Continuous  & \#  Nonzero Coefficient  \\
  \midrule
ex1010-pi & 1468  & 25200& 25200& 0 &0  & 102114  \\
fast0507 & 507 & 63009 & 63009 & 0 & 0 & 409349 \\
ramos3 & 2187 & 2187 &2187&0&0& 32805	 \\
scpj4scip & 1000 & 99947 &99947&0&0& 999893 \\
scpl4 & 2000 & 200000 &200000&0&0& 2000000 \\
% \midrule
dws008-03 & 16344 & 32280&18928&0&13352 & 165168\\
dws008-01 & 6064 & 11096 & 6608 & 0 & 4488  &56400\\
% \midrule
bab2 & 17245 & 147912&147912&0&0 & 2027726\\
bab5 & 4964 & 21600&21600&0&0 & 155520\\
bab6 & 29904 & 114240 &114240&0&0& 1283181\\
neos-3555904-turama & 146493 & 37461 &37461&0&0& 793605 \\
% \midrule
neos-3656078-kumeu & 17656 & 14870 &9755&4455&660& 59292 \\
supportcase17 & 2108 & 	1381&732&235&414 & 5253  \\
fastxgemm-n2r7s4t1 & 6972 & 904 &56&0&848& 22584 \\
% \midrule
tr12-30 & 750 & 1080 &360&0&720& 2508 \\
\bottomrule
\end{tabular}
\end{adjustbox}
\label{table:MIPLIB}
\end{table}

\subsection{Training Details}
Like previous work \citep{PS2023}, our approach maintain the same parameters with \citep{PS2023} without introducing too many extra parameters. We mainly introduce three parameters in CoCo-MILP to control the ranking loss weight, minority variable re-weighting coefficient, and inter-class margin size respectively whose analysis are reported in Appendix E.2. The key parameters in training are summarized in Table \ref{tab:hyperparameters}. 
\begin{table}[h]
\caption{Hyperparameters used in our experiments.}
\centering
%\begin{adjustbox}{width=0.5\textwidth}
{
\begin{tabular}{lcl}
\toprule
  Name &  Value & Description \\
  \midrule
 embed\_size & 64 & The embedding size of the GNN encoder.\\
  num\_epochs & 1000 & Number of max running epochs.\\
 lr & 0.0001 & Learning rate for training.\\
  $\lambda_{\text{rank}}$ & 0.01 & Regularization coefficient, namely the coefficient for Rank Loss.\\
  $\tau$ & 0.1 & The temperature hyperparameter in MSCL.\\
  $\gamma$ & 0.9 & The margin hyperparameter in Rank loss for CA,SC and WA, as for IP, we set $\gamma$ to 0.6\\
\bottomrule
\end{tabular}
%\end{adjustbox}
}
\label{tab:hyperparameters}
\end{table}

\subsection{Inference Details}
To ensure a fair comparison and strong empirical performance, we report the inference-time hyperparameters used by all methods across benchmarks in Table~\ref{tab:inference_hyperparamters}. Each method uses a tuple $(k_0, k_1, \Delta)$, where $k_0$ and $k_1$ determine the top-ranked variable candidates for fixing and unfixing, and $\Delta$ controls the $\ell_1$-norm trust region for solution refinement.
For CoCo-MILP, we perform moderate tuning of these hyperparameters on the validation set of each benchmark to account for the increased complexity and scale of our datasets compared to prior works. While existing baselines such as PS and ConPS provide default hyperparameters in their original papers, we find that these settings often underperform on our harder synthetic and real-world benchmarks. Therefore, we re-tune the parameters for PS and ConPS to ensure their best possible performance under our experimental setup. This tuning procedure significantly improves their feasibility rate and solution quality, and the final configurations are reported in Table~\ref{tab:inference_hyperparamters}.
We also refer readers to our sensitivity analysis in Appendix E.2, where we study the impact of $(k_0, k_1, \Delta)$ on final performance. These results indicate that CoCo-MILP is relatively robust to moderate hyperparameter changes and achieves competitive results across a wide range of values.
All inference experiments are conducted using Gurobi 11.0.3 with a single thread (i.e., \texttt{Threads=1}) and setting the solver focus to solution quality (\texttt{MIPFocus=1}). For experiments involving SCIP, we similarly use single-threaded execution to maintain consistency.

\begin{table}[h]
\caption{Hyperparameters of ($k_0,k_1, \Delta$) used in this paper.}
\centering
\begin{adjustbox}{width=0.7\textwidth}
\begin{tabular}{lccccc}
\toprule
  Benchmark &  CA & SC & IP & WA \\
  \midrule
 PS+Gurobi &  (600,0,1)&(2000,0,100)&(400,5,10)&(0,500,10)\\
 ConPS+Gurobi &  (900,0,50) &(1000,0,200)&(400,5,3)&(0,500,10)\\
 CoCo-MILP+Gurobi & (400, 0, 40) & (1000, 0, 200) & (60, 35, 55) & (20, 200, 100) \\
 PS+SCIP  & (400,0,10) & (2000,0,100)&(400,5,1) &(0,600,5)\\
 ConPS+SCIP  &(900,0,50) &(1000,0,200)&(400,5,3) &(0,400,50)\\
 CoCo-MILP+SCIP & (400, 0, 40) & (1000, 0, 200) & (60, 35, 55) & (20, 200, 100) \\
\bottomrule
\end{tabular}
\end{adjustbox}
\label{tab:inference_hyperparamters}
\end{table}

\section{Additional results}

\subsection{Real-world Datasets}
To further verify the real-world applicability of CoCo-MILP, we evaluate its performance on a selected subset of MIPLIB. Due to the high heterogeneity of this dataset, we follow prior work \citep{liu2025apollomilp} and select a group of structurally similar instances that resemble our synthetic benchmarks.
For each test instance, we construct a training set by identifying approximately five nearby instances based on structural similarity, such as the number of variables, constraints, and sparsity patterns. This instance-wise adaptation enables the model to generalize effectively without the need for global retraining.
During inference, we adopt fixed hyperparameters across all runs: $k_0 = 0.7$, $k_1 = 0.1$, and $\Delta = 800$. These settings are consistent with our main evaluation and are not tuned per instance, highlighting the robustness of CoCo-MILP.
As shown in Table~\ref{table:MIPLIB instance performance}, CoCo-MILP matches or improves upon all baselines on nearly every instance. Notably, it achieves the BKS in all five MIPLIB test cases where the baseline methods fall short. These results confirm that CoCo-MILP maintains strong performance even in challenging real-world scenarios.

\begin{table}[h]
\caption{ The best objectives found by the approaches on each test instance in MIPLIB.
\textit{BKS} represents the best objectives from the website of MIPLIB.}
\vspace{0.25cm}
\centering
\small
\begin{adjustbox}{width=0.9\textwidth}
\begin{tabular}{lcccccc}
\toprule
 &  BKS & Gurobi & PS+Gurobi & ConPS+Gurobi & CoCo-MILP+Gurobi  \\
  \midrule
dws008-03 & 62831.76 & 64452.67 & 71234.06 & 67473.85 & \textbf{66047.78}\\
dws008-01 & 37412.60 & 37412.60 & 	39043.26 & 38817.50 & \textbf{38486.10}\\
% \midrule
bab2 & -357544.31 & -357538.58 & -357449.20 & -357544.31 & \textbf{-357544.31}\\
bab5 & -106411.84 & -106411.84 & -106411.84 & -106411.84 & \textbf{-106411.84}\\
bab6 & -284248.23 & -284248.23 & -284224.56 & -284224.56 & \textbf{-284248.23}\\
neos-3555904-turama & -34.7 & -34.7 & -34.7 & -34.7 & \textbf{-34.7} \\
% \midrule
neos-3656078-kumeu & -13172.2 & -13171.3 & -13114.0 & -13120.6 & \textbf{-13172.20} \\
supportcase17 & 1330.00 & 	1330.00 & 1330.00 & 1330.00 & \textbf{1330.00}  \\
fastxgemm-n2r7s4t1 & 42.00 & 42.00 & 42.00 & 42.00 & \textbf{42.00} \\
% \midrule
tr12-30 & 130596.00 & 130596.00 & 130596.00 & 130596.00 & \textbf{130596.00} \\
\bottomrule
\end{tabular}
\end{adjustbox}
\label{table:MIPLIB instance performance}
\end{table}

\subsection{Hyperparameter Analysis}
To gain a deeper understanding of CoCo-MILP, we conduct a comprehensive analysis of key hyperparameters, including the regularization coefficient $\lambda_{\text{Rank}}$, the ranking margin $\gamma$, the temperature parameter $\alpha$, and the inference-related hyperparameters $k_0$, $k_1$, and $\Delta$.

\paragraph{Regularization Coefficient $\lambda_{\text{Rank}}$}
This coefficient controls the weight of the pairwise ranking loss in the overall contrastive objective. We vary $\lambda_{\text{Rank}}$ in the range $[0.01, 0.08]$ and evaluate its effect on the CA benchmark. As shown in Table~\ref{tab:margin}, increasing the emphasis on inter-class separation does not always improve solution quality. A large coefficient may increase the influence of outliers, such as variables that are correctly classified but located far from their class centers. This can lead to greater inter-class distances while simultaneously reducing overall accuracy. These results suggest that focusing solely on inter-class separation is not sufficient; it is equally important to ensure that samples within the same class are compactly clustered.
\begin{table*}[h]
\centering
\small % 使用较小字体
\caption{Analysis of different regularization coefficients. The ML approaches are implemented using Gurobi, with a time limit set to 1000 seconds. `$\uparrow$' indicates that higher is better, and `$\downarrow$' indicates that lower is better. We mark the \textbf{best values} in bold.}
\label{tab:margin}
\resizebox{0.25\textwidth}{!}{ % 自动缩放至栏宽
\begin{tabular}{@{}lcccccccc@{}}
\toprule
&  \multicolumn{2}{c}{CA  {\scriptsize(BKS 97524.37)}} \\ 
\cmidrule(lr){2-3}
 & Obj $\uparrow$ \hspace{-2mm} & $\text{gap}_{\text{abs}}$ $\downarrow$   \\ 
\midrule
0.01               & \textbf{97338.64}  & \textbf{185.73}  \\
0.03 & 97245.94 &  278.43 \\
0.05 & 97171.55 & 352.82\\
0.08 & 97237.85 & 250.52\\
\bottomrule
\end{tabular}}
\end{table*}

\paragraph{Margin $\gamma$ in Ranking Loss}
The margin parameter $\gamma$ determines the minimum separation enforced between positive and negative variable logits. We observe in Table~\ref{tab:margin} that increasing $\gamma$ from 0.6 to 0.9 steadily improves performance. A larger margin better avoids ambiguous predictions around 0.5, thereby enhancing decision reliability and yielding more accurate final solutions.
\begin{table*}[h]
\centering
\small % 使用较小字体
\caption{Analysis of different margins. The ML approaches are implemented using Gurobi, with a time limit set to 1000 seconds. `$\uparrow$' indicates that higher is better, and `$\downarrow$' indicates that lower is better. We mark the \textbf{best values} in bold.}
\label{tab:margin}
\resizebox{0.25\textwidth}{!}{ % 自动缩放至栏宽
\begin{tabular}{@{}lcccccccc@{}}
\toprule
&  \multicolumn{2}{c}{CA  {\scriptsize(BKS 97524.37)}} \\ 
\cmidrule(lr){2-3}
 & Obj $\uparrow$ \hspace{-2mm} & $\text{gap}_{\text{abs}}$ $\downarrow$   \\ 
\midrule
0.6               & 97264.81  & 259.56  \\
0.7 & 97237.65 &  250.72 \\
0.8 & 97295.39 & 228.72\\
0.9 & \textbf{97338.64}& \textbf{185.73}\\
\bottomrule
\end{tabular}}
\end{table*}

\paragraph{Temperature $\alpha$ in Multi-Sample Contrastive Loss}
The temperature parameter $\alpha$ adjusts the sharpness of the softmax function in the contrastive loss. Smaller values focus the loss more on hard examples. As shown in Table~\ref{tab:temperature}, the best performance is achieved at $\alpha = 0.1$, while both overly low (e.g., 0.05) and high values (e.g., 0.5) degrade solution quality. This confirms that careful tuning of temperature is important to balance hard-negative emphasis and training stability.
\begin{table*}[h]
\centering
\small % 使用较小字体
\caption{Analysis of different temperatures. The ML approaches are implemented using Gurobi, with a time limit set to 1000 seconds. `$\uparrow$' indicates that higher is better, and `$\downarrow$' indicates that lower is better. We mark the \textbf{best values} in bold.}
\label{tab:temperature}
\resizebox{0.25\textwidth}{!}{ % 自动缩放至栏宽
\begin{tabular}{@{}lcccccccc@{}}
\toprule
&  \multicolumn{2}{c}{CA  {\scriptsize(BKS 97524.37)}} \\ 
\cmidrule(lr){2-3}
 & Obj $\downarrow$ \hspace{-2mm} & $\text{gap}_{\text{abs}}$ $\downarrow$   \\ 
\midrule
0.05               & 97216.59  & 0.12  \\
0.1 & \textbf{97338.64} &  \textbf{185.73} \\
0.2 & 97127.90 & 396.47\\
0.3 & 97258.12& 266.25\\
0.5 & 97281.20& 243.17\\
\bottomrule
\end{tabular}}
\end{table*}

\begin{figure}
    \centering
    \includegraphics[width=0.6\linewidth]{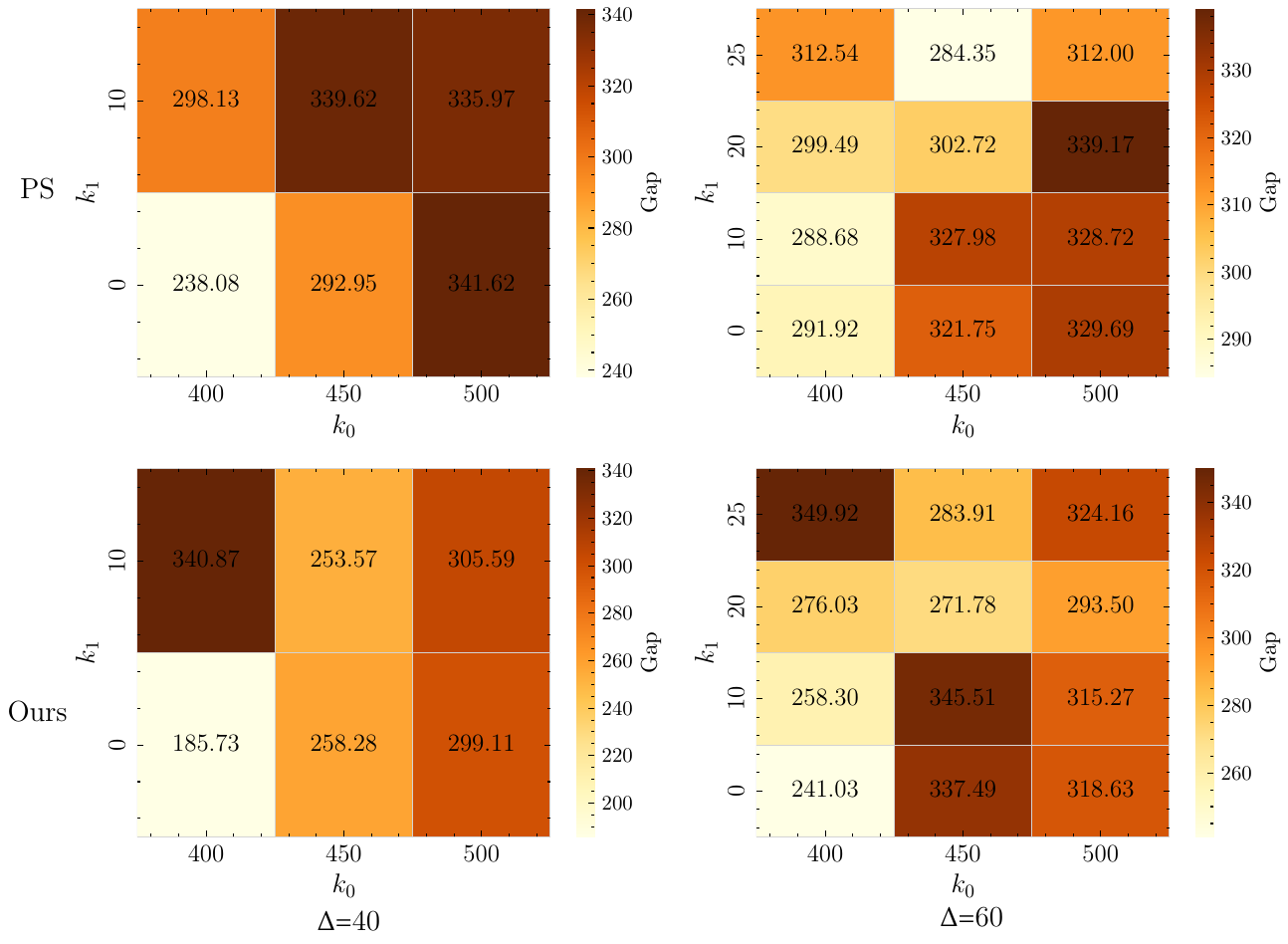}
    \caption{Impact of inference hyperparameters $k_0$, $k_1$, and $\Delta$ on CA benchmark. Each cell reports the average absolute primal gap to BKS under a 1000-second time limit. Upper panels correspond to the PS baseline; lower panels correspond to CoCo-MILP. Smaller values indicate better performance.}
    \label{fig:sensitivity}
\end{figure}

\paragraph{Inference Hyperparameters $k_0$, $k_1$, and $\Delta$}
We further analyze the impact of inference hyperparameters, namely the top-$k_0$ and top-$k_1$ candidate sets for variable selection, and the trust region radius $\Delta$. These parameters jointly determine the search space for the solver after the prediction stage. We evaluate their influence on the CA benchmark by measuring the absolute gap to the BKS under different configurations. The results are visualized as heatmaps in Figure~\ref{fig:sensitivity}, where smaller values indicate better solution quality. Compared to the PS baseline, CoCo-MILP consistently achieves smaller gaps across all settings. Notably, when $\Delta = 40$, our method obtains the best performance at $(k_0=400, k_1=0)$ with a gap of 185.73. This suggests that the predictions produced by CoCo-MILP are already of high quality and require less reliance on extensive neighborhood search. On the other hand, when $\Delta$ increases to 60, the benefit of larger trust regions diminishes for CoCo-MILP, while the baseline shows more sensitivity to this change. Overall, these findings confirm that CoCo-MILP not only improves prediction quality but also stabilizes inference performance under varying search parameters.

\subsection{Evaluation on More Solvers}
To further validate the robustness and generality of CoCo-MILP, we evaluate its performance under the SCIP solver, using the same benchmark setup and a 1000 seconds time limit. The results are summarized in Table \ref{tab:scip-results}. Compared to the vanilla SCIP solver, CoCo-MILP+SCIP consistently improves the primal solution quality across all benchmarks. Specifically, on the SC benchmark, our method reduces the primal gap from 1.48 to 1.21, and on CA, the absolute gap drops by 41.19. For the more challenging IP and WA benchmarks, CoCo-MILP+SCIP achieves notable gap reductions of 0.41 and 0.64, respectively. These results further demonstrate the compatibility of CoCo-MILP with different solvers and its effectiveness as a plug-in predictor, capable of improving solution quality even when integrated into alternative solver backends beyond Gurobi.

\begin{table*}[h]
\centering
\small % 使用较小字体
\caption{Performance comparison on four popular MILP benchmarks using SCIP solver, under a 1000 seconds time limit.
‘$\uparrow$’ indicates higher is better; ‘$\downarrow$’ indicates lower is better.
\textbf{Bold} denotes the best result. All improvements are reported as $\text{gap}_{\text{abs}}$ relative to \textit{BKS}.}
\label{tab:scip-results}
\resizebox{0.85\textwidth}{!}{ % 自动缩放至栏宽
\begin{tabular}{@{}lcccccccccc@{}}
\toprule
& \multicolumn{2}{c}{CA {\scriptsize(BKS 97524.37)}}  & \multicolumn{2}{c}{SC  {\scriptsize(BKS 125.05)}} & \multicolumn{2}{c}{IP {\scriptsize(BKS 11.16)}} &  \multicolumn{2}{c}{WA {\scriptsize(BKS 703.05)}}\\ 
\cmidrule(lr){2-3}
\cmidrule(lr){4-5}
\cmidrule(lr){6-7}
\cmidrule(lr){8-9}
& Obj $\uparrow$& \hspace{-2mm} $\text{gap}_{\text{abs}}$ $\downarrow$ \hspace{-2mm} & Obj $\downarrow$ \hspace{-2mm} & $\text{gap}_{\text{abs}}$ $\downarrow$ \hspace{-2mm} & Obj $\downarrow$  \hspace{-2mm} & $\text{gap}_{\text{abs}} $ $\downarrow$ \hspace{-2mm} & Obj $\downarrow$ \hspace{-2mm} & $\text{gap}_{\text{abs}}$ $\downarrow$          \\ 
\midrule
SCIP                 & 96423 & 1100.47   & 126.43 & 1.48 & 16.55 & 5.39 & 705.77 & 2.72 \\
PS+SCIP              & 96426.46 & 1097.91   & 126.65 & 1.60 & 16.18 & 5.02 & 705.21 & 2.16 \\
ConPS+SCIP           & 96428.83 & 1095.54   & 126.40 & 1.35 & 16.16 & 5.00 & 705.21 & 2.16 \\
CoCo-MILP$+$SCIP & \textbf{96465.09} & \textbf{1059.28} & \textbf{126.26} & \textbf{1.21} & \textbf{16.14} & \textbf{4.98} & \textbf{705.13} & \textbf{2.08}  \\
\bottomrule
\end{tabular}}
\end{table*}

\subsection{Generalization}
To further evaluate the generalization ability of CoCo-MILP, we test its performance on larger CA and SC instances, which contain significantly more constraints and variables, as described in Appendix D.1. We directly apply the model trained on the main evaluation set to these larger instances without any fine-tuning. The results are reported in Table~\ref{tab:generalization_results}, demonstrating that CoCo-MILP exhibits strong generalization capability to larger and more complex problem instances.

\begin{table*}[h]
\centering
\small % 使用较小字体
\caption{Generalization results of 100 larger instances on CA and SC using Gurobi with a 1000 seconds time limit. `$\uparrow$' indicates that higher is better, and `$\downarrow$' indicates that lower is better. We mark the \textbf{best values} in bold.}
\label{tab:generalization_results}
\resizebox{0.5\textwidth}{!}{ % 自动缩放至栏宽
\begin{tabular}{@{}lcccccccccc@{}}
\toprule
& \multicolumn{2}{c}{CA {\scriptsize(BKS 115787.97)}}  & \multicolumn{2}{c}{SC  {\scriptsize(BKS 101.45)}} \\ 
\cmidrule(lr){2-3}
\cmidrule(lr){4-5}
& Obj $\uparrow$& \hspace{-2mm} $\text{gap}_{\text{abs}}$ $\downarrow$ \hspace{-2mm} & Obj $\downarrow$ \hspace{-2mm} & $\text{gap}_{\text{abs}}$ $\downarrow$   \\ 
\midrule
Gurobi                 & 114960.25 & 827.72   & 102.29 & 0.84  \\
PS+Gurobi              & 115228.20 & 559.77   & 102.27 & 0.82  \\
ConPS+Gurobi           & 115343.23 & 444.74   & 102.18 & 0.73  \\
CoCo-MILP$+$Gurobi & \textbf{115463.49} & \textbf{324.48} & \textbf{101.91} & \textbf{0.46}  \\
\bottomrule
\end{tabular}}
\end{table*}

\subsection{Plug-in with other techniques}
To demonstrate the flexibility and broad applicability of CoCo-MILP, it is worth highlighting that our method can be easily integrated with other techniques to achieve enhanced performance. Below, we present two approaches.

\paragraph{Intra-constraint Competition in the Search Phase}
As analyzed in Section 3, modeling intra-constraint competition can significantly improve model performance. A straightforward extension is to apply logit normalization during the search phase, mimicking the effect of ICC layers to amplify the differences among variables within the same constraint. We conduct additional experiments on the SC benchmark, and the results are reported in Table~\ref{tab:competitive_search}. These results demonstrate that incorporating intra-constraint competition into the search phase can lead to further performance improvements.

\begin{table*}[h]
\centering
\small % 使用较小字体
\caption{Results of intra-constraint competition in search phase. The ML approaches are implemented using Gurobi, with a time limit set to 1000 seconds. `$\uparrow$' indicates that higher is better, and `$\downarrow$' indicates that lower is better. We mark the \textbf{best values} in bold.}
\label{tab:competitive_search}
\resizebox{0.5\textwidth}{!}{ % 自动缩放至栏宽
\begin{tabular}{@{}lcccccccc@{}}
\toprule
&  \multicolumn{2}{c}{SC  {\scriptsize(BKS 125.05)}} \\ 
\cmidrule(lr){2-3}
 & Obj $\downarrow$ \hspace{-2mm} & $\text{gap}_{\text{abs}}$ $\downarrow$   \\ 
\midrule
PS+Gurobi               & 125.17 & 0.12  \\
CoCo-MILP+Gurobi           & 125.06 & 0.01  \\
CoCo-MILP+Search Competition$+$Gurobi & \textbf{125.05} & \textbf{0.00}  \\
\bottomrule
\end{tabular}}
\end{table*}

\paragraph{Plug-in with the Stronger Baseline}
As a promising direction in the ML4CO field, we observe that increasingly powerful methods have been proposed, such as Apollo-MILP \citep{liu2025apollomilp} and DiffILO \citep{geng2025differentiable}. Among them, Apollo-MILP is a representative framework that achieves strong performance under the predict-and-search paradigm. It adopts a prediction-and-correction strategy to improve solution quality during the search phase, without altering the training process. The original Apollo-MILP is built upon PS \citep{PS2023}. To evaluate the compatibility of CoCo-MILP, we integrate it into Apollo-MILP by replacing the PS predictor with CoCo-MILP, while keeping the inference hyperparameters unchanged. We conduct experiments on the SC benchmark, and the results are reported in Table~\ref{tab:competitive_search}. When CoCo-MILP is used as a plug-in to enhance Apollo-MILP, the resulting system achieves solutions that match the BKS exactly. This demonstrates the strong complementarity between our method and advanced search frameworks, and further confirms the effectiveness and generality of CoCo-MILP.

\begin{table*}[h]
\centering
\small % 使用较小字体
\caption{Results of CoCo-MILP enhancing with Apollo-MILP. The ML approaches are implemented using Gurobi, with a time limit set to 1000 seconds. `$\uparrow$' indicates that higher is better, and `$\downarrow$' indicates that lower is better. We mark the \textbf{best values} in bold.}
\label{tab:competitive_search}
\resizebox{0.5\textwidth}{!}{ % 自动缩放至栏宽
\begin{tabular}{@{}lcccccccc@{}}
\toprule
&  \multicolumn{2}{c}{SC  {\scriptsize(BKS 125.05)}} \\ 
\cmidrule(lr){2-3}
 & Obj $\downarrow$ \hspace{-2mm} & $\text{gap}_{\text{abs}}$ $\downarrow$   \\ 
\midrule
PS+Gurobi               & 125.17  & 0.12  \\
CoCo-MILP$+$Gurobi & 125.06 &  0.01 \\
Apollo-MILP$+$Gurobi & 125.08 & 0.03\\
CoCo-MILP+Apollo-MILP$+$Gurobi & \textbf{125.05}& \textbf{0.00}\\
\bottomrule
\end{tabular}}
\end{table*}

\end{document}